\documentclass{article}
\usepackage{authblk}

\newcommand{\framework}{\textsc{Echo}} 

\usepackage[preprint]{neurips_2024}

\usepackage{newunicodechar}
\newunicodechar{ }{~}

\usepackage[utf8]{inputenc} 
\usepackage[T1]{fontenc}    
\usepackage{hyperref}       
\usepackage{url}            
\usepackage{booktabs}       
\DeclareUnicodeCharacter{2265}{\geq}
\usepackage{subcaption}    
\usepackage{amsfonts}       
\usepackage{amsmath}        
\usepackage{graphicx}       
\usepackage{enumitem}       
\usepackage{nicefrac}       
\usepackage{microtype}      
\usepackage{xcolor}         

\title{\framework: Decoupling Inference and Training for Large-Scale RL Alignment on Heterogeneous Swarms}

\author[$\ast$1]{Jie Xiao}
\author[$\ast$1]{Changyuan Fan}
\author[1]{Qingnan Ren}
\author[1]{Alfred Long}
\author[1]{Yuchen Zhang}
\author[1]{\authorcr Rymon Yu}
\author[1]{Eric Yang}
\author[1]{Lynn Ai}
\author[2]{Shaoduo Gan}

\affil[1]{\normalsize Gradient Network}
\affil[2]{\normalsize Peking University}

\begin{document}

\maketitle

\def\thefootnote{$\ast$}\footnotetext{Equal Contribution.}

\begin{abstract}
Modern RL-based post-training for large language models (LLMs) co-locate trajectory sampling and policy optimisation on the same GPU cluster, forcing the system to switch between inference and training workloads. This serial context switching violates the single-program–multiple-data (SPMD) assumption underlying today’s distributed training systems. We present \framework, the RL system that cleanly decouples these two phases across heterogeneous “inference” and “training” swarms while preserving statistical efficiency. Echo introduces two lightweight synchronization protocols: a sequential pull mode that refreshes policy weights according to API call for minimal bias, and an asynchronous push–pull mode that streams version-tagged rollouts through a replay buffer to maximise hardware utilisation. Training four representative RL workloads with Qwen3-4B, Qwen2.5-7B, Qwen3-30B-A3B-Thinking-2507, and Qwen3-32B on a geographically distributed cluster, \framework\ matches a fully co-located \textsc{Verl} baseline in convergence speed and final reward while off-loading trajectory generation to commodity edge hardware. These promising results demonstrate that large-scale RL for LLMs could achieve datacentre-grade performance using decentralised, heterogeneous resources.

\end{abstract}


\section{Introduction}

Reinforcement learning (RL) has become the
\emph{de facto} recipe for aligning billion-parameter language models
with human preferences, such as RLHF\cite{ouyang2022training}, RLAIF\cite{lee2023rlaif}, DPO\cite{rafailov2023direct} and their variants.
All of these algorithms share a multi-program,
multi-data (MPMD) execution pattern:
\emph{(i)} an \textbf{inference phase} that generates trajectories under
the current policy, followed by
\emph{(ii)} a \textbf{training phase} that converts those trajectories
into gradient updates.
State-of-the-art systems, e.g.,\ DeepSpeed\,RLHF\cite{yao2023deepspeed} and \textsc{Verl}\cite{sheng2024hybridflow}, run the two phases on the \emph{same}
high-bandwidth cluster, implicitly synchronising through global barriers
and expensive NVLink / InfiniBand fabrics. Since trajectory sampling and policy optimization have very different hardware footprints, running both phases on a single hardware pool forces the cluster to serially contextual-switch between two different workloads. This back-and-forth violates the single-program–multiple-data (SPMD) assumption that underpins most large-scale training systems, hindering utilisation and scalability.

The natural remedy is to \emph{split}
trajectory generation and policy optimization onto
distinct, heterogeneous swarms.
However, naive decoupling may break the implicit guarantees provided
by co-location: rollouts may be generated by stale policies; learners may optimise with
out-of-date data; and, as a result, the entire training can diverge.

In this paper, we introduce \textsc{Echo}, the RL system that
solves the above problem through a pair of synchronization mechanisms, as shown in Figure~\ref{fig:system}. The \emph{sequential} mode has trainers \emph{pull} rollouts via an API, refreshing policy weights beforehand to mirror classic PPO and minimise bias.
The \emph{asynchronous} mode lets samplers \emph{push} tagged rollouts into a replay buffer while a coordinator keeps tracking version drift, maximising device utilisation.

On the inference side, we build upon \textsc{Parallax}, a fully
decentralised pipeline-parallel engine developped by Gradient Network that converts geographically
dispersed consumer GPUs and Apple-Silicon devices into a single,
high-throughput sampler via dynamic KV-cache management and continuous
batching.
On the training side we adopt the community-standard
\textsc{Verl} stack, adding robust support for
LoRA-style parameter-efficient finetuning.

We train four typical RL workloads from scratch on \textbf{Qwen3-4B},
\textbf{Qwen2.5-7B}, \textbf{Qwen3-30B-A3B-Thinking-2507}, and \textbf{Qwen3-32B}—using both \framework\ and a fully co-located \textsc{Verl} baseline. In every case, \framework\ matches the baseline’s convergence speed and final reward while off-loading trajectory generation to a pool of heterogeneous edge devices. These results demonstrate that large-scale RL can exploit decentralized, commodity hardware without sacrificing training efficiency.

\paragraph{Contributions.}
\begin{enumerate}[leftmargin=*]
  \item We formulate the information-synchronisation challenge arising
        from decoupling RL trajectory generation and policy optimisation.
  \item We design two complementary protocols, i.e., \emph{Sequential} and
        \emph{Asynchronous}, that bound policy lag without datacentre-class
        interconnects.
  \item We implement \framework, featuring the \textsc{Parallax} inference
        engine and an enhanced \textsc{Verl} trainer with LoRA support.
  \item Our extensive experiments show that a
        heterogeneous, geographically distributed cluster can train
        LLMs via RL at parity with high-end datacenter baselines.
\end{enumerate}


\section{System Design}

\subsection{System Overview}
\label{sec:system-overview}

Conventional RL stacks co-locate trajectory generation and policy
updates on the same high-bandwidth cluster, implicitly relying on
barrier-style synchronisation to guarantee that every gradient step
consumes up-to-date data.  
\framework\ deliberately \emph{splits} these two phases
onto distinct, heterogeneous swarms—\underline{an \textit{inference swarm}}
that samples trajectories and \underline{a \textit{training swarm}} that
optimises the policy—so that each can run on the hardware best suited to
its workload.  
The price of this flexibility is a non-trivial \textbf{information synchronisation problem}: trajectory freshness and weight staleness must
be controlled explicitly; otherwise, the decoupled system may diverge or
converge more slowly than monolithic baselines.  
Tackling this challenge requires a careful analysis of the
\emph{data-dependency chain} between policy parameters, sampled
trajectories, and gradient updates.

\begin{figure}[h]
  \centering
  \includegraphics[width=0.85\linewidth]{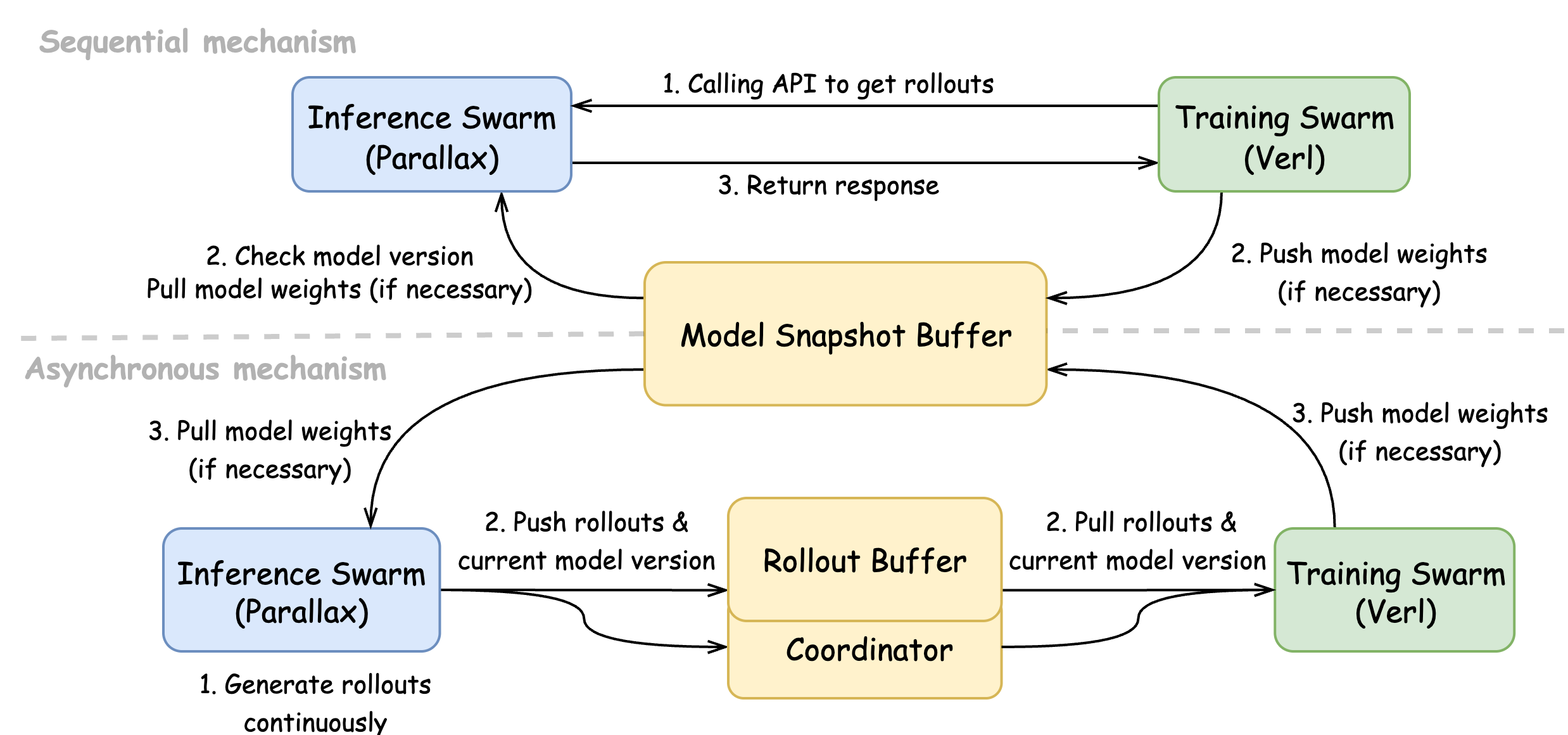}
  \caption{\framework\ architecture and the two synchronisation
           mechanisms between the training and inference swarms.}
  \label{fig:system}
\end{figure}

As shown in Figure~\ref{fig:system}, the \framework\ system is composed of two complementary protocols focusing on different optimisation
goals:

\begin{itemize}[leftmargin=*]
  \item \textbf{Sequential mechanism (accuracy-centric).}
        The training swarm \emph{pulls} trajectories on demand via an
        API, and the inference peer first checks the version of its local weights before producing the rollouts. Therefore, the version of the policy weights used to generate rollouts can be precisely controlled. This path delivers the lowest bias and therefore the best \emph{statistical accuracy}.
  \item \textbf{Asynchronous mechanism (efficiency-centric).}
        The inference swarm streams rollouts continuously to a shared
        \texttt{rollout buffer}, tagging each sample with the parameter
        version. The training swarm consumes these version-stamped
        batches at its own pace and requests a weight refresh only when
        the observed version drift exceeds a threshold. This path maximises
        \emph{hardware utilisation} and is preferred when abundant,
        under-used inference capacity is available.
\end{itemize}

Both protocols adopt \texttt{model-snapshot buffer} that stores
the model checkpoints to be synchronized. Together \framework\ let users trade a small
amount of statistical bias for a large gain in throughput, or vice
versa, without altering the rest of the system. 

Beyond cross-swarm synchronisation, \framework\ embeds state-of-the-art optimisations inside each swarm. In the Inference side, We build upon \textsc{Parallax}, a comprehensive distributed inference framework that runs LLMs seamlessly across heterogeneous devices—from Nvidia GPUs to Apple-Silicon laptops. The training side extends the community-standard \textsc{Verl}\,RL stack, preserving broad algorithmic compatibility (PPO, GRPO, DPO, \dots) while adding first-class support for parameter-efficient adapters such as LoRA.

The details of each component of \framework\ are described in the following sections.

\subsection{Synchronization Mechanism}

Decoupling rollout generation from gradient calculation transforms the classical, single‑loop RL pipeline into two semi‑independent yet mutually‑constrained swarms. In this section, we describe how \textbf{(i)} trajectories flow \emph{upward} from the inference swarm to the training swarm and \textbf{(ii)} policy snapshots flow \emph{downward} in the opposite direction, while keeping the correctness of the end-to-end training computation.

\subsubsection{Rollouts}
\label{sec:rollouts}

Our framework exposes two complementary data paths from the inference swarm back to the trainers.  

\paragraph{1) Inference‑as‑an‑API (trainer‑driven, pull‑based).}
In this mode the inference swarm surfaces a \emph{trajectory‑generation API}.  
The training swarm invokes this endpoint whenever it needs fresh data, thereby \textbf{explicitly} deciding \emph{what} to sample, \emph{when} to sample, and whether the sampler should refresh its local model weights before serving the request.  

\begin{itemize}[leftmargin=*]
\item \textbf{API contract.}  
  The endpoint follows an \emph{OpenAI‑compatible} JSON schema so that existing tooling (e.g.\ OpenAI Python SDK) can be reused:  
\begin{verbatim}
POST /v1/rl/trajectories
{
  "model": "ECHO-Infer",
  "param_version": "theta_e",   // trainer's current weights
  "prompts": [p_1, …, p_k],
  "algo": "PPO"                 // or PPO‑variant identifier
}
→
{
  "param_version": "theta_e",   // echoed back for traceability
  "trajectories": [
    { "tokens": [...],
      "actions": [...],
      "logprobs": [...],
      "values": [...],
      "rewards": [...],
      "terminal": true },
    …
  ]
}
\end{verbatim}
  Every trajectory yields exactly the $(s,a,\log\pi_\theta(a|s),v_\theta(s),r)$
  fields required by PPO and its popular variants.

\item \textbf{Call logic.}  
  From the trainer’s perspective the call is \emph{blocking}: optimisation does not proceed until the requested batch returns.  
  The simplest policy is to invoke the API once per optimisation step, but more sophisticated schedulers can batch multiple steps or interleave asynchronous prefetchers.  
  On the inference side the request handler (i) compares its local weights against the caller’s \texttt{param\_version}; if the version gap exceeds a pre‑configured threshold it reloads the newer weights, otherwise it continues with the current checkpoint; (ii) executes the rollout script to produce $k$ trajectories; (iii) streams the batch back to the caller.

\item \textbf{Applicable scenarios.}  
  This pull‑based path is ideal when the algorithm demands \emph{fine‑grained, runtime control} over both the data distribution and the policy version—e.g.\ active‑learning curricula, on‑policy off‑diagonal exploration, or methods with low tolerance to policy staleness.  
  Because the semantics match those of a single‑process PPO loop, researchers can port centrally‑designed RL code without rewriting dataflow logic.  
  The blocking nature of the call makes it best suited to short rollouts or compute‑rich inference swarms where latency is dominated by a single forward pass rather than network transfers.
\end{itemize}

\paragraph{2) Replay‑Buffer Streaming (asynchronous, push–pull).}
In this regime the inference and training swarms progress \emph{concurrently} rather than in a strict master‑worker loop.  
The inference swarm continuously produces trajectories with its current policy and pushes them into a distributed replay buffer, while the training swarm pulls mini‑batches from that buffer at its own pace.  
No direct dependency exists between the two loops except for co‑ordinated weight synchronisations.

\begin{itemize}[leftmargin=*]
\item \textbf{Overlapping the rollout generation and the training.}  
  Once optimisation starts, every inference node autonomously samples the environment with its resident policy, annotates each rollout with a \texttt{param\_version} tag, and appends the resulting batch to the replay buffer.  
  Meanwhile the training swarm streams mini‑batches from the buffer and updates the policy in parallel, so rollout generation and gradient computation overlap in time, amortising end‑to‑end latency.

\item \textbf{Coordinator.}  
  Because the strong synchronization is intentionally broken, a lightweight coordinator orchestrates two critical invariants.  
  First, it balances throughput by aligning the inference batch size to an integer multiple of the trainer’s mini‑batch size, mitigating “batch fragmentation’’ in the replay buffer.  
  Second, it monitors the version skew between the policy used to \emph{produce} a batch and the one used to \emph{consume} it; when the skew exceeds a configurable threshold, it broadcasts a \texttt{sync\_weight} to both sides.

\item \textbf{Applicable scenarios.}  
  This path suits algorithms that tolerate limited policy staleness—e.g.\ experience replay variants of PPO or off‑policy critics—because generated trajectories may lag one or more updates behind the current actor.  
  It is particularly advantageous when individual trajectories involve long‑horizon or multi‑turn inference, where overlapping compute can mask inference latency and yield a substantial utilisation improvement.
\end{itemize}

\subsubsection{Model Weights}

In a centralised PPO loop, the \emph{same} policy is used for rollout generation and gradient updates, so actor–learner staleness is naturally bounded. After decoupling the two swarms, however, the convergence efficiency highly depends on keeping their weight versions \emph{sufficiently} aligned. Whenever synchronisation is required, the trainer serialises its current checkpoint, uploads it to \texttt{model-snapshot buffer}, and the inference swarm fetches and activates the new policy.

\begin{itemize}[leftmargin=*]
\item \textbf{API‑triggered synchronisation.}  
  In the \emph{Inference‑as‑an‑API} path the trainer
  embeds its latest \texttt{param\_version} in every request.  
  Upon receipt the inference node compares this tag with the version of
  its resident weights; if the numeric or hash distance exceeds a
  pre‑configured threshold, it starts to download the new checkpoint and reloads the latest policy, and only then serves the rollout
  request.

\item \textbf{Coordinator‑triggered synchronization (replay‑buffer
      mode).}  
  For the asynchronous replay‑buffer path, the coordinator continually tracks two counters: the learner’s global
  update step $t_{\text{train}}$ and the current update step
  used by the inference swarm $t_{\text{infer}}$.  
  When $t_{\text{train}}-t_{\text{infer}}>\Delta_{\max}$, the
  coordinator broadcasts a \texttt{sync\_weight} command. This scheme amortises transfer cost over many trajectories while bounding the maximum policy lag to $\Delta_{\max}$. Note that pre-defining a max staleness degree is just the simplest strategy. By leveraging the \texttt{Coordinator}, more complicated synchronization algorithms can be easily integrated into \framework, such as comparing the distribution distance of the model output, or the statistic features of two models, and so on. We leave this in the future work.
\end{itemize}

\subsection{Fully Decentralized Inference Engine - \textsc{Parallax}}

\textsc{Parallax} links a pool of geographically dispersed, consumer‑grade
accelerators, like 40/50‑series GPUs, Apple M‑series, and so on, into a \emph{single pipeline‑parallel sampler}.  Unlike datacenter engines that
rely on RDMA fabrics and homogeneous hardware, \textsc{Parallax} assumes
nothing beyond commodity Ethernet and heterogeneous devices.

\paragraph{Layer Allocation.}
Here we take a simple yet effective algorithm to allocate the model layers among devices. During initialisation the inference engine probes every node for its peak
FLOPS ${f_{\text{peak}}}$ and free memory ${m_{\text{free}}}$.  It then partitions
the $L$ transformer layers into $S$ \textbf{static stages}
$\{\mathcal{L}_1,\dots,\mathcal{L}_S\}$ that satisfy
\[
  \forall s,\quad
  T_{\mathrm{comp}}(\mathcal{L}_s)\;\approx\;
  \frac{1}{S}\sum_{j=1}^{L}T_{\mathrm{comp}}(\ell_j),
\]
so each stage is expected to finish a forward step in roughly the same time.
Because this balancing happens \emph{once} before roll‑outs start, the runtime
needs no layer migration, credit rebalancing, or token stealing—keeping the
wire protocol thin enough for ordinary 10–25 GbE links.

\subsection{Training with Algorithm Diversity and Hardware Efficiency}

The \textsc{Echo} training swarm consumes trajectory batches, applies a
chosen RL algorithm, and performs gradient updates on the policy
parameters.  
Large‑language models impose substantial compute and memory pressure, so
the training cluster is provisioned with high‑bandwidth GPU nodes (e.g.\
NVLink‑connected A100/H100 pods) and exploits a hybrid of data
parallelism, pipeline parallelism, and tensor parallelism to scale both
model and batch size efficiently.  
By decoupling training from rollout generation we can separately tune
parallel strategy, memory budget, and GPU topology on each side—freeing
the training swarm to pursue maximal throughput without constraining the
latency‑oriented inference swarm.

\paragraph{1) RL Algorithm Diversity.}
LLM‑centric reinforcement learning is a rapidly evolving research area,
with continual proposals for new credit‑assignment schemes, preference
optimisers, and reward models.  
Compatibility with the SOTA algorithms is therefore an important
design principle.  
Our framework inherits the \emph{modular} trainer stack of
\textsc{Verl}, one of the community’s most widely adopted RL libraries
for language models.  
Because \textsc{Verl} factorises its optimiser, advantage estimator, and reward
inference into interchangeable components, any algorithm implemented in
\textsc{Verl}, such as vanilla PPO, KL‑constrained PPO, GRPO, DPO, or emerging
variants, can be integrated into \textsc{Echo} with negligible glue code;
the only requirement is that the optimiser exposes a
\texttt{step()} interface consuming mini‑batches drawn from the shared
buffer.

\paragraph{2) Parameter‑efficient Training.}
Low‑rank adaptation (LoRA) has become an indispensable technique for
LLM post‑training, reducing effective parameter count by two orders of
magnitude and slashing wall‑clock time as well as energy cost.  
In a decentralised setting these savings translate directly into lower
checkpoint sizes and thus smaller over‑the‑wire synchronisation cost.  
While the upstream \textsc{Verl} repository offers only experimental LoRA
support, we contribute a production‑ready extension that transparently
wraps \textbf{FSDP} and \textbf{Megatron} back‑ends.

\section{Evaluation}

\begin{figure*}[t]
  \centering
  \begin{subfigure}[t]{\linewidth}
    \centering
    \includegraphics[width=0.55\linewidth]{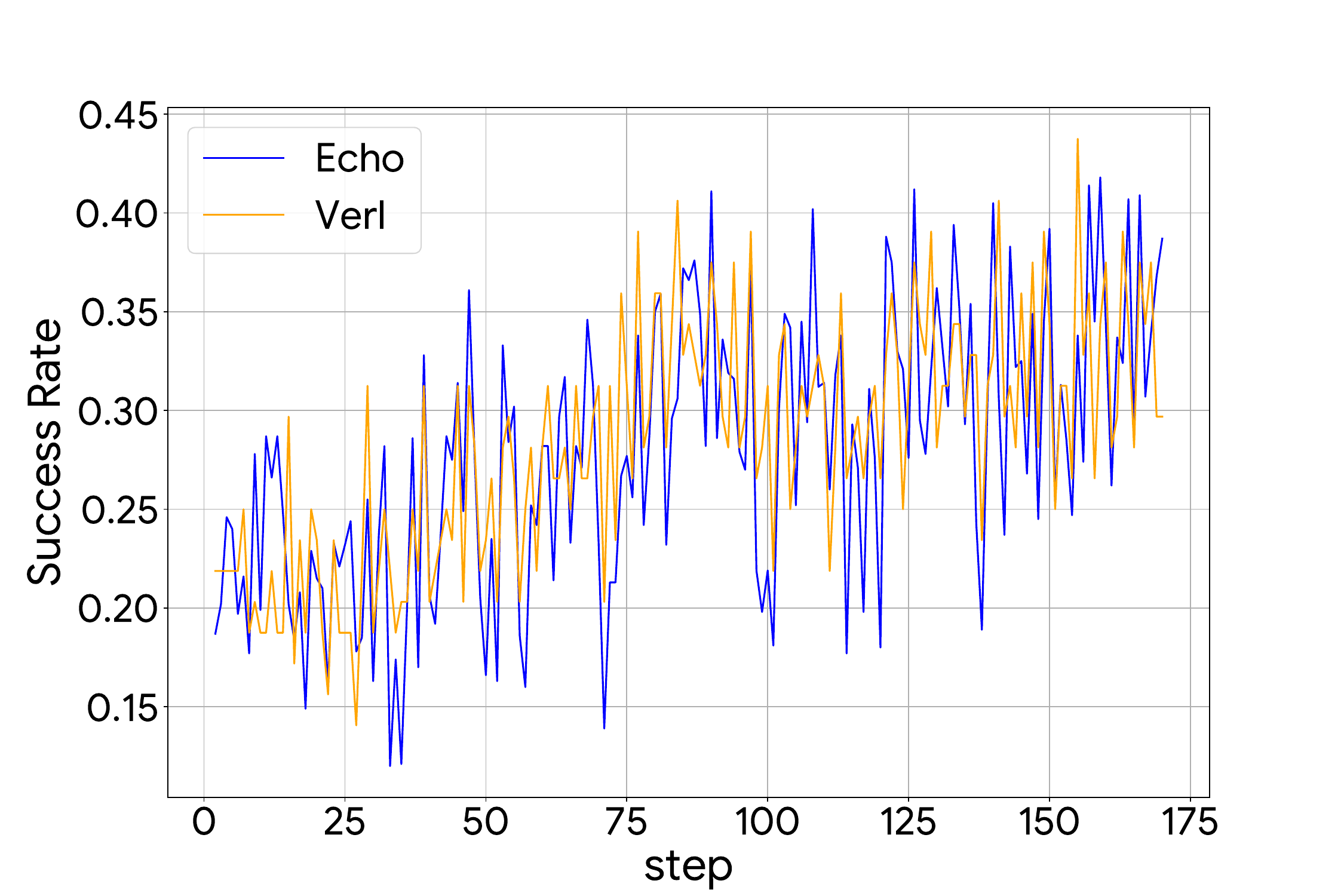}
    \caption{Sokoban w/ Qwen3-4B}
    \label{fig:RAGEN_4b}
  \end{subfigure}
  \begin{subfigure}[t]{\linewidth}
    \centering
    \includegraphics[width=0.55\linewidth]{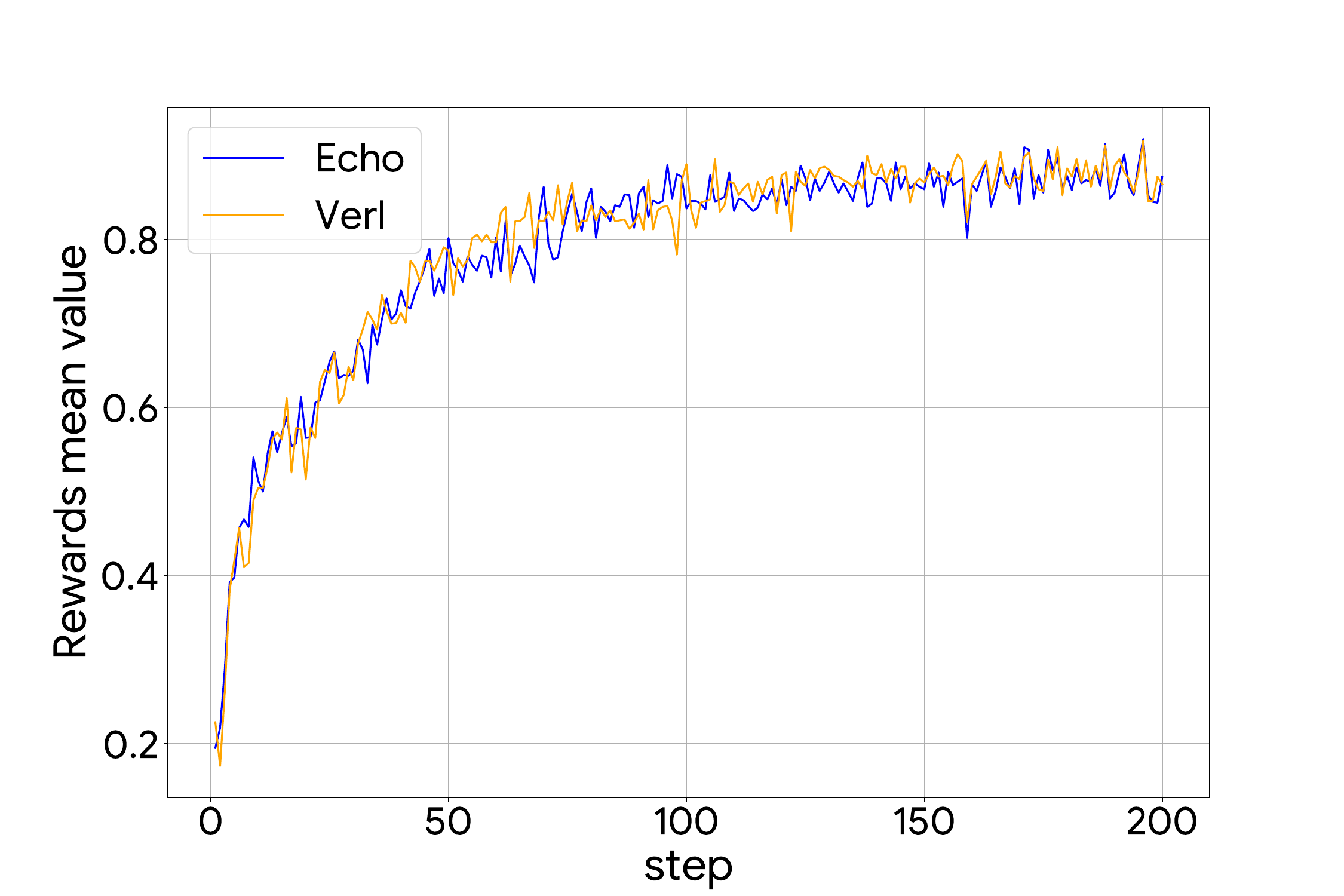}
    \caption{Math w/ Qwen2.5-7B}
    \label{fig:math_7b}
  \end{subfigure}
  \begin{subfigure}[t]{\linewidth}
    \centering
    \includegraphics[width=0.55  \linewidth]{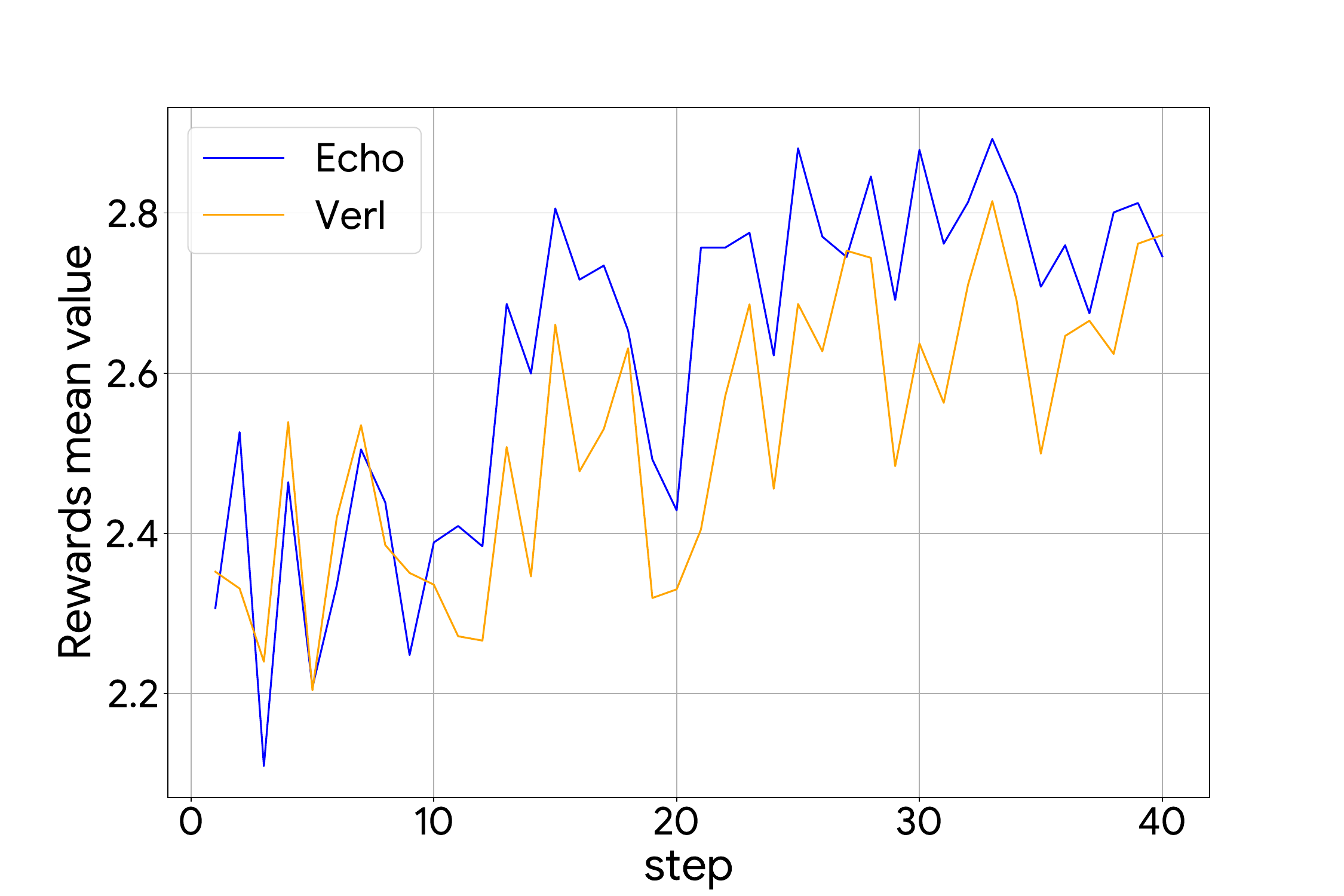}
    \caption{K\&K w/ Qwen3-32B}
    \label{fig:KK_32B}
  \end{subfigure}
  \caption{Training curves on the three tasks. \framework\ always matches the centralized baseline in convergence speed
           and final return.}
  \label{fig:training_curves}
\end{figure*}

The central question we seek to answer by experiments is simple yet crucial:  
\emph{Can a decoupled architecture—where trajectory generation and
policy optimisation run on separate, heterogeneous swarms—match the
training speed and final performance of a conventional, fully
co-located RL stack?}  

To this end, we reproduce a suite of canonical RL workloads that have
well-established learning curves under a datacentre setup
and re-train them under \framework.

\subsection{Experiment Setup }

We conduct experiments on three representative tasks using three model scales from the Qwen series. The primary objective is to compare the performance of the \textsc{Verl} and \framework\ frameworks across different models and tasks.

\paragraph{Models. }
We utilize the open-source Qwen series models as base models for training: Qwen3-4B, Qwen2.5-7B, Qwen3-32B and Qwen3-30B-A3B-Thinking-2507.

\paragraph{Tasks and Datasets. }

\begin{figure}[t]

\centering

\includegraphics[width=0.6\linewidth]{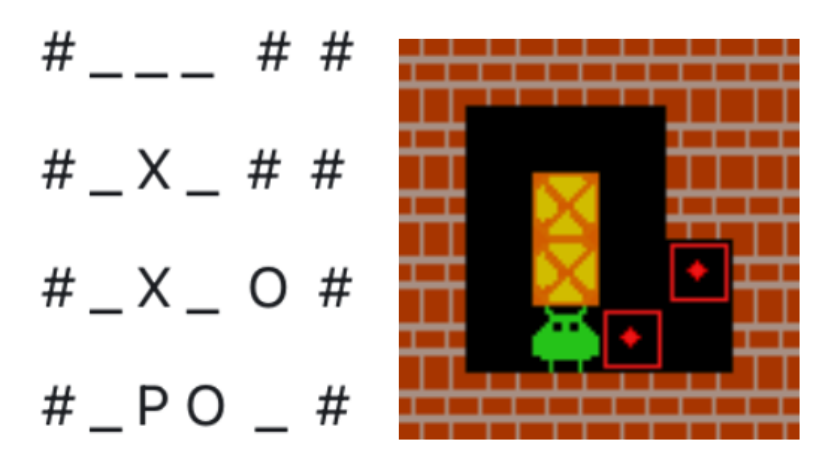}
\caption{Sokoban Environment}
\label{fig:sokoban_gym}

\end{figure}

Sokoban \cite{junghanns2001sokoban} : We utilize the Sokoban puzzle (Figure~\ref{fig:sokoban_gym}) to study multi-turn agent interaction in environments requiring irreversible long-term planning. In Sokoban, the agent must push boxes to designated goals within a grid world under constrained steps. Crucially, boxes can only be pushed, not pulled back, demanding significant foresight to avoid irreversible dead ends. We implement this using a Gym\cite{brockman2016openaigym} environment configured with a 6 × 6 grid and 2 boxes. The reward structure promotes efficiency and accuracy: +1 for each box correctly placed on a target, -1 for each box off-target, +10 upon full task completion, and -0.1 per action penalty.

Mathematical Problem Solving: Focuses on verifiable mathematical problems. We use the Eurus-2-RL-Math dataset \cite{cui2025processreinforcementimplicitrewards} for training. Model performance is evaluated on the following test sets: MATH500 \cite{hendrycks2021measuringmathematicalproblemsolving} , AIME 2024 \cite{li2024numinamath} , AMC \cite{li2024numinamath}, AIME 2025, OlympiadBench \cite{he2024olympiadbench}, and Minerva \cite{lewkowycz2022solving}.

Knights and Knaves (K\&K) Puzzles \cite{xie2025memorizationlargelanguagemodels}: This dataset consists of algorithmically generated reasoning puzzles. Characters are either Knights (always truthful) or Knaves (always lying). The goal is to determine each character's type based on their statements.

\paragraph{Hardware.}

\textsc{Verl} Framework: Utilizes one machine with 8~\(\times\)~A100 80GB GPUs for all tasks.

\framework\ Framework: On the inference side, we use a distributed network of 6 nodes (\textsc{Parallax} ): 3~\(\times\)~RTX 5090 and 3~\(\times\)~Mac M4 Pro . On the training side, we use a server equipped with 4~\(\times\)~A100 GPUs.

\paragraph{Training Recipe.}

RL training employs the GRPO algorithm under both the \textsc{Verl} and \framework\ reinforcement learning frameworks. Hyperparameter settings are task-specific:

Sokoban (Base model: Qwen3-4B): Learning Rate: 1e-6, Batch Size: 6, Rollout N: 16, KL Coefficient: 0.001.

Sokoban (Base model: Qwen3-30B-A3B-Thinking-2507): Learning Rate: 1e-6, Batch Size: 8, Rollout N: 16, KL Coefficient: 0.001.

Mathematical Problem Solving (Base model: Qwen2.5-7B): Learning Rate: 5e-7, Batch Size: 256, Rollout N: 8, KL Coefficient: 0.0.

K\&K  (Base model: Qwen3-32B, trained with LoRA): Learning Rate: 1e-6, Batch Size: 64, Rollout N: 8, KL Coefficient: 0.001, LoRA rank: 64, LoRA alpha: 32.

\begin{table}[t!]
\caption{Model performance on Sokoban task}
\label{tab:success-rate}
\centering

\begin{tabular}{lc}
\hline
\textbf{Model} & \textbf{Success Rate (\%)} \\
\hline
Qwen3-4B & 21.8 \\
Qwen3-4B-\framework(GRPO) & \textbf{34.0} \\
Qwen3-30B-A3B-Thinking-2507 & 72.75 \\
Qwen3-30B-A3B-Thinking-2507-\framework(GRPO) & \textbf{82.20} \\
Deepseek-R1 & 75.75 \\
Qwen3-235B-A22B-Thinking-2507 & 79.68 \\
gpt-oss-120b & 79.69 \\
\hline
\end{tabular}

\end{table}

\begin{table}[t]
\caption{Model performance on math reasoning tasks. For AIME and AMC, the results are avg.@32}
\label{tab:math-performance}
\centering
\resizebox{\textwidth}{!}{
\begin{tabular}{lccccccc}
\toprule
\textbf{Model} & \textbf{AIME24} & \textbf{AIME25} & \textbf{AMC} & \textbf{MATH-500} & \textbf{OlympiadBench} & \textbf{Minerva} & \textbf{Avg.} \\
\midrule
Qwen2.5-7B & 2.7\% & 1.9\% & 22.0\% & 44.6\% & 19.7\% & 20.9\% & 18.6\% \\
Qwen2.5-32B & 5.3\% & 2.1\% & 27.9\% & 62.4\% & 25.4\% & 33.5\% & 26.1\% \\
Qwen2.5-7B-\framework(GRPO) & \textbf{13.1\%} & \textbf{6.9\%} & \textbf{45.6\%} & \textbf{75.4\%} & \textbf{37.0\%} & \textbf{50.7\%} & \textbf{38.1\%} \\
\bottomrule
\end{tabular}
}
\end{table}

\begin{table}[h!]

\caption{Model performance on K\&K logic puzzle task across different degrees of difficulty}
\label{tab:accuracy-tasks}
\centering
\begin{tabular}{lccccccc}
\hline
\textbf{Model} & \textbf{2} & \textbf{3} & \textbf{4} & \textbf{5} & \textbf{6} & \textbf{7} & \textbf{8} \\
\hline
Qwen3-32B & 0.98 & 0.99 & 0.98 & 0.99 & 0.98 & 0.96 & 0.95 \\
Deepseek-R1 & \textbf{1.00} & 0.97 & 0.95 & 0.93 & 0.91 & 0.93 & 0.91 \\
o3-mini-high & \textbf{1.00} & \textbf{1.00} & \textbf{1.00} & \textbf{1.00} & \textbf{0.99} & 0.98 & 0.98 \\
o4-mini & \textbf{1.00} & \textbf{1.00} & 0.96 & 0.94 & 0.97 & 0.93 & 0.87 \\
Qwen3-32B-\framework (GRPO w/ Lora) & 0.99 & \textbf{1.00} & \textbf{1.00} & \textbf{1.00} & \textbf{0.99} & \textbf{1.00} & \textbf{0.99} \\
\hline
\end{tabular}

\end{table}

In the \framework\ framework, we employ the Sequential mechanism (accuracy-centric) for now, such that we can precisely verify the correctness of the implementation of \framework. Specifically, the model is synchronized in each inference step during rollout generation. This setup ensures the training process logically remains fully aligned with \textsc{Verl} throughout the experiments. The tuning of more-advanced synchronization algorithms is left for future work.

\subsection{Results}

\paragraph{End-to-End Comparison of Training Convergence.}
Figure~\ref{fig:training_curves} shows the training curves of three RL tasks specified in the above section. In every plot, we contrast the canonical, fully co-located \textsc{Verl} baseline with our decoupled implementation running on \framework. In the Figure~\ref{fig:RAGEN_4b}, success rates fluctuate heavily by design, yet the blue and orange traces track each other throughout training. After around 160 steps, both variants stabilise around a 0.40 success rate, again indicating no degradation in final performance. In the Figure~\ref{fig:math_7b}, the two curves are almost indistinguishable: they accelerate rapidly to 0.8 in the first 80 steps, then converge smoothly toward the 0.9 ceiling. Wall-clock measurements show the same number of actor steps to reach 0.85, validating that our synchronisation protocol maintains sample efficiency. In the Figure~\ref{fig:KK_32B}, we scale our experiment model to 32B. Both systems begin around a mean critic reward of 2.3 and rise steadily. From step 10 onwards, the decoupled variant even edges ahead, finishing at $\sim$2.8 versus the baseline’s $\sim$2.75. The overlap in the early phase and the matching plateau confirm that separating rollout generation does not introduce optimization bias.

\paragraph{Performance Improvement by \framework.}
In the Sokoban task, we have post-trained 2 models: Qwen3-4B and Qwen3-30B-A3B-Thinking-2507. As shown in Table~\ref{tab:success-rate}, Qwen3-4B-\framework\ surpassed the Qwen3-4B by $13\% $. Noteably, Qwen3-30B-A3B-Thinking-2507-\framework\ even achieves the best performance over DeepSeek-R1, Qwen3-235B-A22B-Thinking-2507, and gpt-oss-120b. In math reasoning tasks, we post-trained Qwen2.5-7B with \framework\ on Eurus-2-RL-Math dataset\cite{cui2025processreinforcementimplicitrewards}, and it outperformed the Qwen2.5-32B on all six test datasets, achieving a $12\% $ improvement on average, as shown in Table~\ref{tab:math-performance}. Our experiments on the K\&K task show that training under \framework\ significantly improves model performance, especially in more difficult multi-person scenarios (6-8 people). The Qwen3-32B model fine-tuned with GRPO and LoRA achieved near-perfect scores (all metrics $\geq$ 0.99), surpassing o3-mini-high, DeepSeek-R1 and o4-mini, achieving the best performance. See Table~\ref{tab:accuracy-tasks} for detailed results.


\section{Conclusion and Future Work}
We presented \textsc{Echo}, a dual-swarm RL system that disentangles
trajectory sampling from policy optimisation so each phase can exploit
the hardware that suits it best.
Echo combines (i) a fully decentralised, pipeline-parallel inference
engine (\textsc{Parallax}) capable of harnessing heterogeneous edge
devices and (ii) an enhanced \textsc{Verl} trainer that retains broad
algorithm support and production-grade LoRA fine-tuning.
Two lightweight synchronisation protocols—\emph{sequential} for
accuracy and \emph{asynchronous} for efficiency—bound policy lag without
requiring datacentre-class interconnects and can be swapped with only
minimal orchestration changes.
Experiments on Qwen3-4B, Qwen2.5-7B, Qwen3-30B-A3B-Thinking-2507, and Qwen3-32B show that Echo
matches a fully co-located baseline in convergence speed and final
reward while off-loading trajectory generation to commodity edge
hardware, demonstrating the viability of large-scale RL on
geographically dispersed resources.

\paragraph{Future Work.}
The dominant cost in our current deployment is model-parameter
synchronisation.  We plan to attack this bottleneck on two fronts:

\begin{enumerate}[leftmargin=*]
    \item \textbf{Reducing synchronisation frequency.}
          We will design \emph{runtime-adaptive} sync policies that use
          training-time statistics, e.g.,\ KL divergence between successive
          policies, gradient-norm trends, or replay-buffer skew, to decide
          \emph{whether and when} the inference swarm should refresh its
          weights, thereby avoiding unnecessary transfers.
    \item \textbf{Reducing synchronisation volume.}
          We intend to explore compression and quantisation techniques
          tailored to single-direction snapshot distribution, such as
          low-precision LoRA deltas, sparsity-aware encoding, or
          entropy-constrained chunking, to shrink each weight push by an
          order of magnitude.
\end{enumerate}

Together, adaptive frequency control and communication-efficient weight
encoding will pave the way for deploying Echo on even wider classes of
edge devices, thereby unlocking a truly global, decentralised pool of RL
compute.

\newpage

\bibliographystyle{abbrv}
\bibliography{references} 

\begin{thebibliography}{10}

\bibitem{brockman2016openaigym}
G.~Brockman, V.~Cheung, L.~Pettersson, J.~Schneider, J.~Schulman, J.~Tang, and W.~Zaremba.
\newblock Openai gym, 2016.

\bibitem{cui2025processreinforcementimplicitrewards}
G.~Cui, L.~Yuan, Z.~Wang, H.~Wang, W.~Li, B.~He, Y.~Fan, T.~Yu, Q.~Xu, W.~Chen, J.~Yuan, H.~Chen, K.~Zhang, X.~Lv, S.~Wang, Y.~Yao, X.~Han, H.~Peng, Y.~Cheng, Z.~Liu, M.~Sun, B.~Zhou, and N.~Ding.
\newblock Process reinforcement through implicit rewards, 2025.

\bibitem{he2024olympiadbench}
C.~He, R.~Luo, Y.~Bai, S.~Hu, Z.~Thai, J.~Shen, J.~Hu, X.~Han, Y.~Huang, Y.~Zhang, J.~Liu, L.~Qi, Z.~Liu, and M.~Sun.
\newblock Olympiadbench: A challenging benchmark for promoting agi with olympiad-level bilingual multimodal scientific problems.
\newblock In L.-W. Ku, A.~Martins, and V.~Srikumar, editors, {\em Proceedings of the 62nd Annual Meeting of the Association for Computational Linguistics (Volume 1: Long Papers)}, pages 3828--3850, Bangkok, Thailand, 2024. Association for Computational Linguistics.

\bibitem{hendrycks2021measuringmathematicalproblemsolving}
D.~Hendrycks, C.~Burns, S.~Kadavath, A.~Arora, S.~Basart, E.~Tang, D.~Song, and J.~Steinhardt.
\newblock Measuring mathematical problem solving with the math dataset, 2021.

\bibitem{junghanns2001sokoban}
A.~Junghanns and J.~Schaeffer.
\newblock Sokoban: Enhancing general single-agent search methods using domain knowledge.
\newblock {\em Artificial Intelligence}, 129(1):219--251, 2001.

\bibitem{lee2023rlaif}
H.~Lee, S.~Phatale, H.~Mansoor, K.~R. Lu, T.~Mesnard, J.~Ferret, C.~Bishop, E.~Hall, V.~Carbune, and A.~Rastogi.
\newblock Rlaif: Scaling reinforcement learning from human feedback with ai feedback.
\newblock 2023.

\bibitem{lewkowycz2022solving}
A.~Lewkowycz, A.~Andreassen, D.~Dohan, E.~Dyer, H.~Michalewski, V.~Ramasesh, A.~Slone, C.~Anil, I.~Schlag, T.~Gutman-Solo, et~al.
\newblock Solving quantitative reasoning problems with language models.
\newblock In {\em Advances in Neural Information Processing Systems}, volume~35, pages 3843--3857. Curran Associates, Inc., 2022.

\bibitem{li2024numinamath}
J.~Li, E.~Beeching, L.~Tunstall, B.~Lipkin, R.~Soletskyi, S.~Huang, K.~Rasul, L.~Yu, A.~Q. Jiang, Z.~Shen, et~al.
\newblock {Numinamath}: The largest public dataset in ai4maths with 860k pairs of competition math problems and solutions.
\newblock \url{https://huggingface.co/datasets/numinamath/numinamath}, 2024.
\newblock Hugging Face repository, 13:9.

\bibitem{ouyang2022training}
L.~Ouyang, J.~Wu, X.~Jiang, D.~Almeida, C.~Wainwright, P.~Mishkin, C.~Zhang, S.~Agarwal, K.~Slama, A.~Ray, et~al.
\newblock Training language models to follow instructions with human feedback.
\newblock {\em Advances in neural information processing systems}, 35:27730--27744, 2022.

\bibitem{rafailov2023direct}
R.~Rafailov, A.~Sharma, E.~Mitchell, C.~D. Manning, S.~Ermon, and C.~Finn.
\newblock Direct preference optimization: Your language model is secretly a reward model.
\newblock {\em Advances in neural information processing systems}, 36:53728--53741, 2023.

\bibitem{sheng2024hybridflow}
G.~Sheng, C.~Zhang, Z.~Ye, X.~Wu, W.~Zhang, R.~Zhang, Y.~Peng, H.~Lin, and C.~Wu.
\newblock Hybridflow: A flexible and efficient rlhf framework.
\newblock {\em arXiv preprint arXiv: 2409.19256}, 2024.

\bibitem{xie2025memorizationlargelanguagemodels}
C.~Xie, Y.~Huang, C.~Zhang, D.~Yu, X.~Chen, B.~Y. Lin, B.~Li, B.~Ghazi, and R.~Kumar.
\newblock On memorization of large language models in logical reasoning, 2025.

\bibitem{yao2023deepspeed}
Z.~Yao, R.~Y. Aminabadi, O.~Ruwase, S.~Rajbhandari, X.~Wu, A.~A. Awan, J.~Rasley, M.~Zhang, C.~Li, C.~Holmes, et~al.
\newblock Deepspeed-chat: Easy, fast and affordable rlhf training of chatgpt-like models at all scales.
\newblock {\em arXiv preprint arXiv:2308.01320}, 2023.

\end{thebibliography}
\end{document}